\documentclass[letterpaper, 10 pt, conference]{ieeeconf}  

\IEEEoverridecommandlockouts                              

\usepackage{cite}
\usepackage{amsmath,amssymb,amsfonts}
\usepackage{algorithmic}
\usepackage{graphicx}
\usepackage{textcomp}
\usepackage{xcolor}
\usepackage{csquotes}
\usepackage{caption}
\usepackage{subcaption}
\usepackage{hyperref}
\usepackage{multirow}
\usepackage{amsmath} 
\usepackage{comment}

\title{Federated Learning for Predicting Mild Cognitive Impairment to Dementia Conversion\\

\thanks{This research has been supported by grants 346934, 358944 (Flagship of Advanced Mathematics for Sensing Imaging and Modeling) from the Research Council of Finland; grant 351849 from the Research Council of Finland under the frame of ERA PerMed (“Pattern-Cog”), and it has been conducted as part of the ITEA Secur-e-Health project funded by Business Finland (Joint Action ID 4220/31/2021).}
}

\author{Gaurang Sharma$^{1,2}$, Elaheh Moradi$^{1}$, Juha Pajula$^{2}$, Mika Hilvo$^{2}$, and Jussi Tohka$^{1}$, \\ for the Alzheimer’s Disease Neuroimaging Initiative$^{*}$
\thanks{*Data used in preparation of this article were obtained from the Alzheimer's Disease Neuroimaging Initiative (ADNI) database (adni.loni.usc.edu). As such, the investigators within the ADNI contributed to the design and implementation of ADNI and/or provided data but did not participate in analysis or writing of this report. A complete listing of ADNI investigators can be found at: \url{http://adni.loni.usc.edu/wp-content/uploads/how_to_apply/ADNI_Acknowledgement_List.pdf}}
\thanks{$^{1}$A.I. Virtanen Institute for Molecular Sciences, University of Eastern Finland, Kuopio, Finland
        {\tt\small Email: firstname.lastname@uef.fi}}%
\thanks{$^{2}$VTT Technical Research Centre of Finland Ltd,
        Espoo, Finland
        {\tt\small Email: firstname.lastname@vtt.fi}}%
}

\begin{document}

\maketitle
\thispagestyle{empty}
\pagestyle{empty}

\begin{abstract}

Dementia is a progressive condition that impairs an individual's cognitive health and daily functioning, with mild cognitive impairment (MCI) often serving as its precursor. 
The prediction of MCI-to-dementia conversion has been well studied, but previous studies have almost always focused on traditional Machine Learning (ML)—based methods that require sharing sensitive clinical information to train predictive models. This study proposes a privacy-enhancing solution using Federated Learning (FL) to train predictive models for MCI-to-dementia conversion without sharing sensitive data, leveraging socio-demographic and cognitive measures. We simulated and compared two network architectures, Peer-to-Peer (P2P) and client-server, to enable collaborative learning. Our results demonstrated that FL had comparable predictive performance to centralized ML, and each clinical site showed similar performance without sharing local data. Moreover, the predictive performance of FL models was superior to site-specific models trained without collaboration.  This work highlights that FL can eliminate the need for data sharing without compromising model efficacy.

\end{abstract}

\section{Introduction}
Neurodegeneration refers to the gradual and progressive loss of the structure and functions of the neurons, resulting in a diverse group of disorders such as Alzheimer's disease, Parkinson's disease and others. Neurodegenerative diseases cause a decrease in cognitive functions, affecting memory and/or behavioral abilities, finally interfering with the quality of life and daily activities of an individual. At this stage, the condition is considered dementia \cite{arvanitakis2019diagnosis}. A key concept is Mild Cognitive Impairment (MCI), which is the in-between stage between typical cognitive skills and dementia. Invididuals with MCI are at a greater risk of developing dementia, but in many cases, the symptoms of MCI stay the same or even improve.

\begin{figure}[t]
  \begin{subfigure}[]{0.235\textwidth}
    \includegraphics[width=\textwidth]{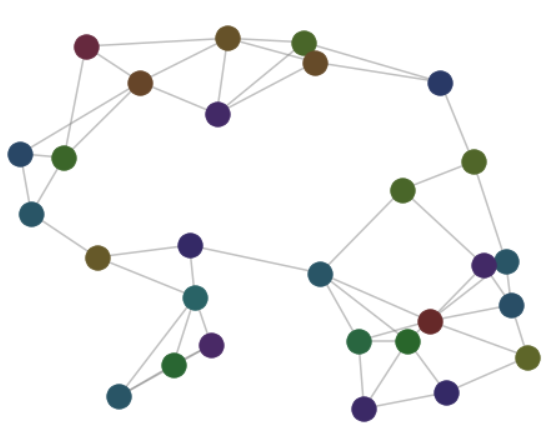}
  \end{subfigure}
  \hfill
  \begin{subfigure}[]{0.235\textwidth}
    \includegraphics[width=\textwidth]{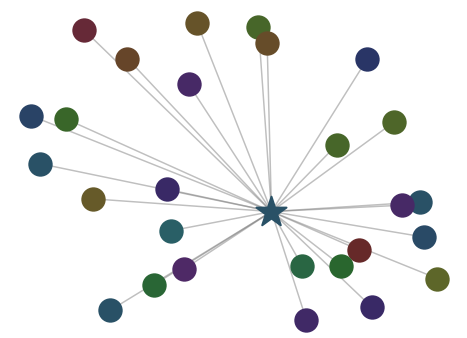}
  \end{subfigure}
  \caption{Empirical graphs visualizing the FL architectures proposed to maintain data privacy: P2P (left panel) and client-server (right panel). Nodes represent the ADNI sites, and edges enable the connectivity among nodes. In client-server architecture, the star sign represents the server or computational node.}
  \label{fig:graphs}
\end{figure}

\begin{figure*}[t] 
    \centering
    \begin{subfigure}[t]{0.32\textwidth}
        \centering
        \includegraphics[width=\linewidth]{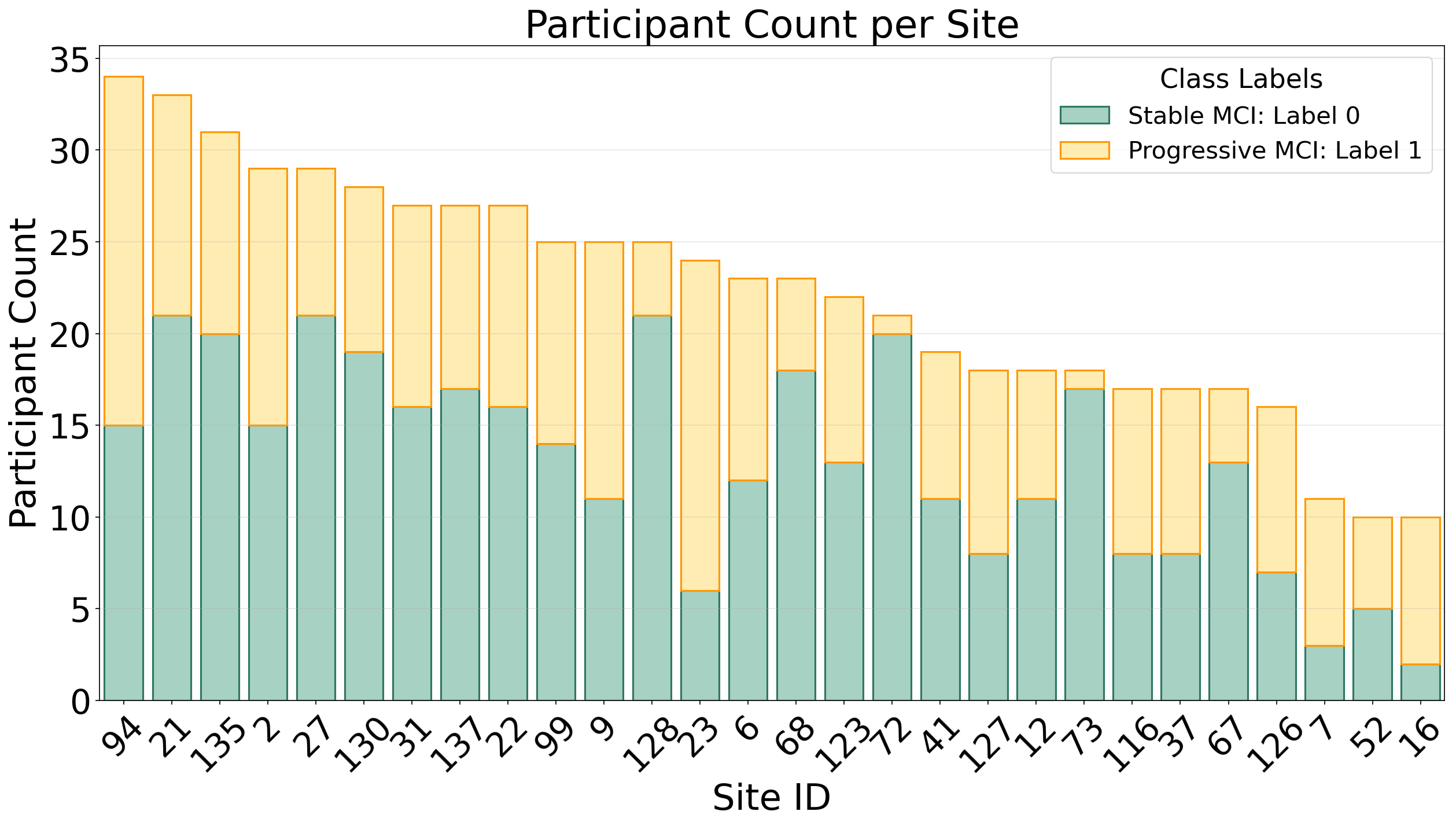} 
        \label{fig:data_3}
    \end{subfigure}
    \hfill
    \begin{subfigure}[t]{0.32\textwidth}
        \centering
        \includegraphics[width=\linewidth]{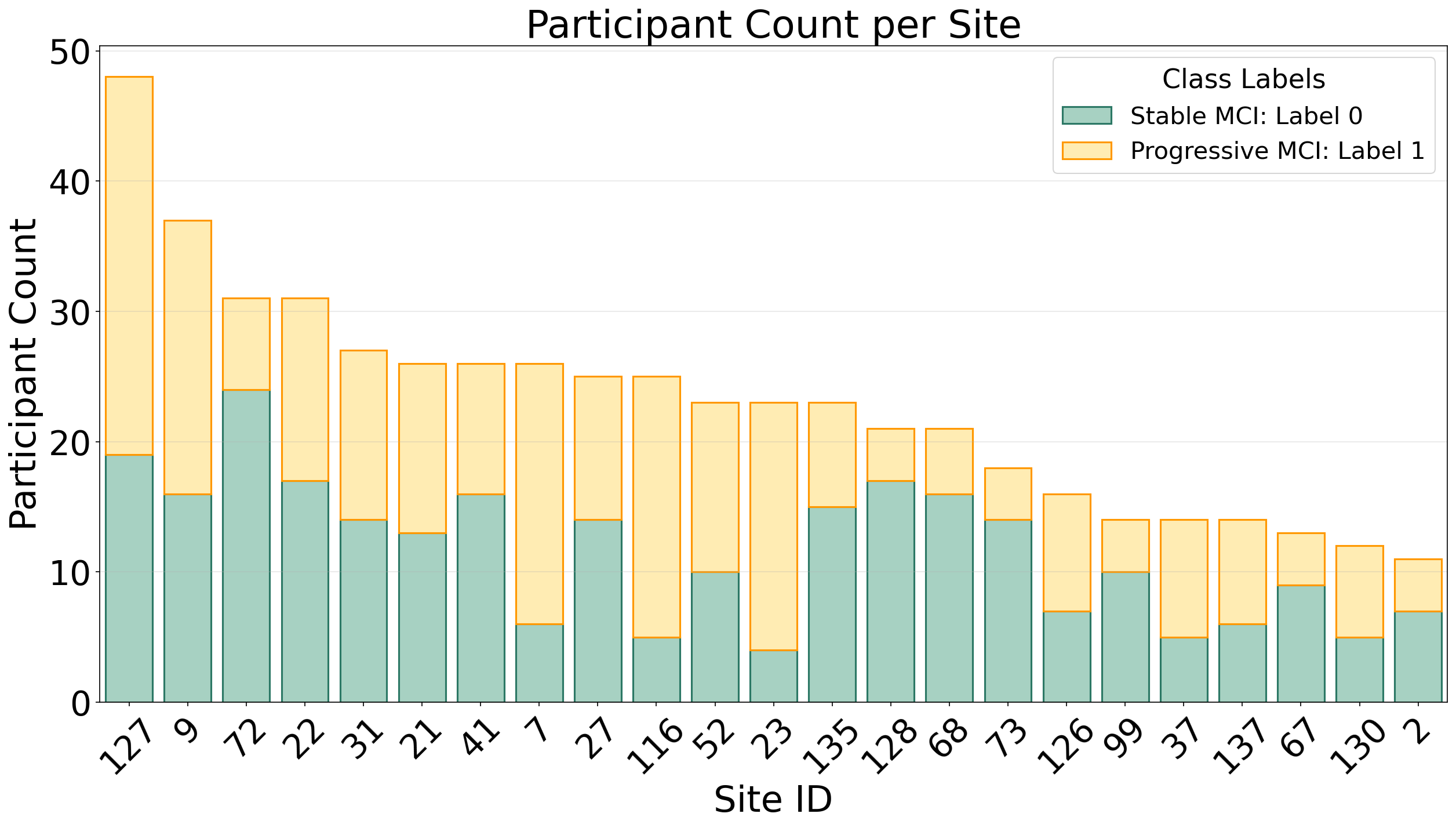} 
        \label{fig:data_4}
    \end{subfigure}
    \hfill
    \begin{subfigure}[t]{0.32\textwidth}
        \centering
        \includegraphics[width=\linewidth]{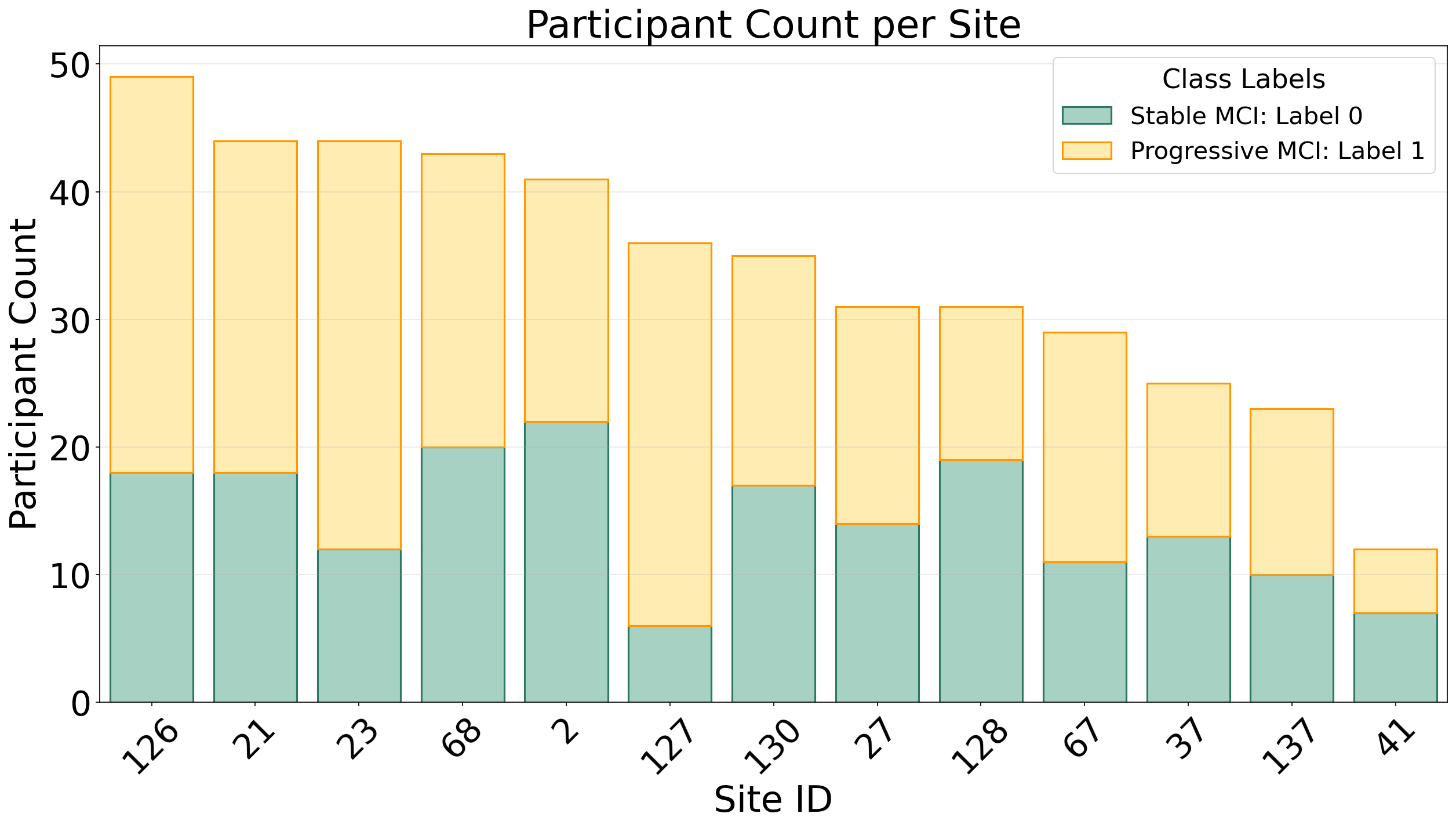} 
        \label{fig:data_5}
    \end{subfigure}
    \caption{Data distribution at ADNI sites based on the clinical status of an individual, where sMCI (label 0) and pMCI (label 1) are visualized with green and orange bars, respectively. From left, 3, 4, and 5 years of follow-up.   
    }
    \label{fig:data}
\end{figure*}



An early dementia diagnosis is essential for guiding appropriate management strategies and implementing timely interventions \cite{watson2018timely}, given the increased evidence of successful dementia risk reduction and prevention actions \cite{frisoni2023dementia} and medication to slow down the cognitive decline \cite{jamainternmed}. Predicting whether an individual suffering from MCI will have a dementia diagnosis in future has been considered to be a key aspect towards early dementia diagnosis and large-scale studies on this MCI-to-dementia conversion prediction are clearly warranted.


The Machine Learning (ML) based prediction of MCI-to-dementia conversion has been well studied \cite{ chen2022prediction, ANSART2021101848}, with cognitive test-based models proving to be highly effective \cite{Modat2023}. 
Cognitive tests are important tools, typically acquired in clinical settings at the initial stage of cognitive assessment. They are cost-effective and do not require expensive equipment, making them accessible for widespread use.
A conventional approach for training MCI-to-dementia prediction models using cognitive assessment data is centralized ML, where data from all sites is aggregated at a central location for model training \cite{colliot2023machine}.
However, aggregating data from several sources poses significant ethical and legal problems as well as the possibility of data leaks and unauthorized access \cite{ateniese2015hacking, yigzaw2022health, lamas2015ethical, minssen2020eu}. Thus, an alternative approach is to retain data locally and develop models collaboratively, enhancing privacy. 

Federated Learning (FL), a privacy-enhancing technology, offers the practice of keeping data private for data controllers and training a predictive model collaboratively.  McMahan et al. \cite{mcmahan2017communication} proposed the FedAvg algorithm for decentralized training without sharing raw data. Since then, numerous optimization strategies, architectures, and application-specific adaptations have emerged \cite{shanmugam2023federated}. Researchers have focused on communication efficiency, handling data heterogeneity, and integrating privacy-enhancing techniques \cite{kairouz2021advancesopenproblemsfederated, 9084352} and considered health care related applications \cite{pfitzner2021federated}.

In dementia-related research, FL has been applied to address privacy concerns and enable collaborative analysis with most works focusing on neuroimaging and biomarker data. Lei et al. applied FL with transformers to adapt multi-site neuroimaging data \cite{lei2023federated}. Huang et al. introduced FedCM, a privacy-preserving framework utilizing MRI and 3D-CNNs for Alzheimer's disease detection \cite{9630382}. Stripelis et al. developed a secure FL approach using fully homomorphic encryption using MR images \cite{stripelis2021secure, stripelis2023secureprivatefederated}. Elsersy et al. leveraged blood biomarkers for early dementia detection \cite{elsersy2023federated}, and Mitrovska et al. integrated FL with Secure Aggregation (SecAgg) for Alzheimer's detection using structural MRI (sMRI) data\cite{mitrovska2024secure}. However, modalities used in these previous FL works are either expensive, invasive, or not readily available \cite{chouliaras2023use, dulewicz2022biomarkers}. 
Conversely, cognitive tests are non-invasive, cost-effective, very widely available, and often more sensitive to early cognitive changes \cite{michalowsky2017cost, turner2023standardized, borland2022clinically, elkana2015sensitivity}. However, despite their advantages, cognitive assessments remain largely unexplored within FL frameworks in dementia research, especially for predicting dementia progression or conversion, representing a significant research gap.
Moreover, the previous studies have primarily focused on client-server based setups and have not examined Peer-to-Peer (P2P) architectures.

This paper fills these gaps by presenting the FL methodology for MCI-to-dementia conversion, which predicts cognitive status after 3, 4, and 5 years from the baseline using cognitive and socio-demographic features. 
We evaluate how FL methods perform compared to traditional centralized ML, examine the impact of collaborative parameters on site-specific model performance, and compare different FL methods (client-server and P2P, see Fig. \ref{fig:graphs}) on overall performance, generalization, and computational efficiency.


\section{Data}
\label{sec:data}

The data used was obtained from the ADNI (\url{http://adni.loni.usc.edu}). The ADNI was launched in 2003 as a public-private partnership. 
The primary goal of the cohort has been to test whether serial MRI, PET, other biological markers, as well as clinical and neuropsychological assessments, can be combined to measure the progression of MCI and early AD. For up-to-date information, see \url{http://www.adni-info.org}.

We included participants from all the phases of ADNI whose initial diagnosis was MCI and who had baseline demographics, APOE4, cognitive test results, and sufficient follow-up information available. The list of participants IDs is available at \enquote{\url{https://github.com/GaurangSharma18/FL-MCI-Dementia}}. 
We used socio-demographic and cognitive data at the baseline visit as predictors. The socio-demographic data included age, sex, education level, and  Apolipoprotein E4 (ApoE4) genotype. Cognitive features encompassed a variety of standardized test scores of a participant obtained to assess cognitive function and memory. These features were Clinical Dementia Rating - Sum of Boxes (CDR-SB), Alzheimer's Disease Assessment Scale - Cognitive Subscale (ADAS-Cog13), ADAS Delayed Word Recall (ADASQ4), Mini-Mental State Examination (MMSE), and several aspects of the Rey Auditory Verbal Learning Test (RAVLT) (Immediate recall, Learning, Forgetting, and Percent Forgetting). Moreover, Logical Memory—Delayed Recall, Trail Making Test Part B (TRABSCOR), and Functional Activities Questionnaire (FAQ) were used as cognitive predictors. Both cognitive and socio-demographic features were obtained from the ADNIMERGE table. More details on the cognitive test procedures and clinical assessment are available in \cite{petersen2010alzheimer} and the procedures manual \footnote{ADNI procedures Manual: \url{https://adni.loni.usc.edu/wp-content/uploads/2024/02/ADNI3_Procedures_Manual_v3.0.pdf}}.


To train the model for predicting future cognitive status using the data at the baseline visit, participant's diagnosis at the end of the follow-up period served as the ground truth. Based on the final diagnosis, participants were assigned to one of two groups: progressive MCI (pMCI), for those who received dementia diagnoses in their last two follow-up assessments, or stable MCI (sMCI), for those who consistently maintained an MCI diagnosis throughout the entire follow-up.

We formulated three case studies with different follow-up times, see Table \ref{tab:data}. The cohort contains data from 63 sites. 
However, we merged the sites with fewer than ten participants with other sites. As a result, the site count was reduced to 28, 23, and 13 for cases 1, 2, and 3, respectively. The participant count per site and case is visualized in Fig. \ref{fig:data}.

\begin{table}[]
\resizebox{\columnwidth}{!}{%
\begin{tabular}{|l|l|l|l|l|}
\hline
\textbf{Case} & \textbf{Follow-up Period} & \textbf{Sites} & \textbf{pMCI} & \textbf{sMCI} \\ \hline
\textbf{1}    & 3 years                   & 28             & 256           & 368           \\ \hline
\textbf{2}    & 4 years                   & 23             & 256           & 269           \\ \hline
\textbf{3}    & 5 years                   & 13             & 256           & 187           \\ \hline
\end{tabular}%
}
\caption{The number of participants per followup period and conversion status. The site counts shown reflect the number of sites after merging those with fewer than ten participants.}
\label{tab:data}
\end{table}

\begin{table*}[ht]
\centering
\resizebox{\textwidth}{!}{
\begin{tabular}{|lllllll|}
\hline
\multicolumn{1}{|l|}{\textbf{Methodology}} &
  \multicolumn{1}{l|}{\textbf{Framework}} &
  \multicolumn{1}{l|}{\textbf{Optimizer}} &
  \multicolumn{1}{l|}{\textbf{ROC AUC Score}} &
  \multicolumn{1}{l|}{\textbf{Balanced Accuracy}} &
  \multicolumn{1}{l|}{\textbf{Specificity}} &
  \textbf{Sensitivity} \\ \hline
\multicolumn{7}{|c|}{\textbf{Case 1: 3 years from the baseline}} \\ \hline
\multicolumn{1}{|l|}{\textbf{Centralized ML}} &
  \multicolumn{1}{l|}{Logistic Regression} &
  \multicolumn{1}{l|}{Gradient Descent} &
  \multicolumn{1}{l|}{0.903 ± 0.057} &
  \multicolumn{1}{l|}{0.842 ± 0.050} &
  \multicolumn{1}{l|}{0.911 ± 0.053} &
  0.774 ± 0.071 \\ \hline
\multicolumn{1}{|l|}{\multirow{2}{*}{\textbf{Federated}}} &
  \multicolumn{1}{l|}{P2P} &
  \multicolumn{1}{l|}{FedGD} &
  \multicolumn{1}{l|}{0.891 ± 0.050} &
  \multicolumn{1}{l|}{0.821 ± 0.072} &
  \multicolumn{1}{l|}{0.880 ± 0.075} &
  0.762 ± 0.082 \\ \cline{2-7} 
\multicolumn{1}{|l|}{} &
  \multicolumn{1}{l|}{Client-server} &
  \multicolumn{1}{l|}{FedAvg} &
  \multicolumn{1}{l|}{0.890 ± 0.049} &
  \multicolumn{1}{l|}{0.822 ± 0.073} &
  \multicolumn{1}{l|}{0.878 ± 0.077} &
  0.765 ± 0.083 \\ \hline
\multicolumn{7}{|c|}{\textbf{Case 2: 4 years from the baseline}} \\ \hline
\multicolumn{1}{|l|}{\textbf{Centralized ML}} &
  \multicolumn{1}{l|}{Logistic Regression} &
  \multicolumn{1}{l|}{Gradient Descent} &
  \multicolumn{1}{l|}{0.920 ± 0.055} &
  \multicolumn{1}{l|}{0.860 ± 0.049} &
  \multicolumn{1}{l|}{0.896 ± 0.055} &
  0.825 ± 0.055 \\ \hline
\multicolumn{1}{|l|}{\multirow{2}{*}{\textbf{Federated}}} &
  \multicolumn{1}{l|}{P2P} &
  \multicolumn{1}{l|}{FedGD} &
  \multicolumn{1}{l|}{0.919 ± 0.054} &
  \multicolumn{1}{l|}{0.859 ± 0.047} &
  \multicolumn{1}{l|}{0.896 ± 0.054} &
  0.822 ± 0.064 \\ \cline{2-7} 
\multicolumn{1}{|l|}{} &
  \multicolumn{1}{l|}{Client-server} &
  \multicolumn{1}{l|}{FedAvg} &
  \multicolumn{1}{l|}{0.919 ± 0.054} &
  \multicolumn{1}{l|}{0.861 ± 0.051} &
  \multicolumn{1}{l|}{0.899 ± 0.063} &
  0.824 ± 0.064 \\ \hline
\multicolumn{7}{|c|}{\textbf{Case 3: 5 years from the baseline}} \\ \hline
\multicolumn{1}{|l|}{\textbf{Centralized ML}} &
  \multicolumn{1}{l|}{Logistic Regression} &
  \multicolumn{1}{l|}{Gradient Descent} &
  \multicolumn{1}{l|}{0.926 ± 0.052} &
  \multicolumn{1}{l|}{0.856 ± 0.073} &
  \multicolumn{1}{l|}{0.856 ± 0.095} &
  0.856 ± 0.067 \\ \hline
\multicolumn{1}{|l|}{\multirow{2}{*}{\textbf{Federated}}} &
  \multicolumn{1}{l|}{P2P} &
  \multicolumn{1}{l|}{FedGD} &
  \multicolumn{1}{l|}{0.927 ± 0.045} &
  \multicolumn{1}{l|}{0.854 ± 0.065} &
  \multicolumn{1}{l|}{0.849 ± 0.089} &
  0.859 ± 0.055 \\ \cline{2-7} 
\multicolumn{1}{|l|}{} &
  \multicolumn{1}{l|}{Client-server} &
  \multicolumn{1}{l|}{FedAvg} &
  \multicolumn{1}{l|}{0.927 ± 0.045} &
  \multicolumn{1}{l|}{0.855 ± 0.066} &
  \multicolumn{1}{l|}{0.852 ± 0.090} &
  0.859 ± 0.055 \\ \hline
\end{tabular}}
\caption{
Performance comparison of FL and Centralized ML across all sites. The FL methodology used FedGD in P2P and FedAvg in client-server architectures for the collaborative training. The listed values are the performance mean$\pm$ standard deviation across the 10 folds on a common test set.
}
\label{tab:logisticRegression}
\end{table*}

\begin{figure*}[t]  
    \centering
    \begin{subfigure}[c]{0.5\linewidth}  
        \centering
        \vfill  
        \includegraphics[height=12cm,width=\linewidth]{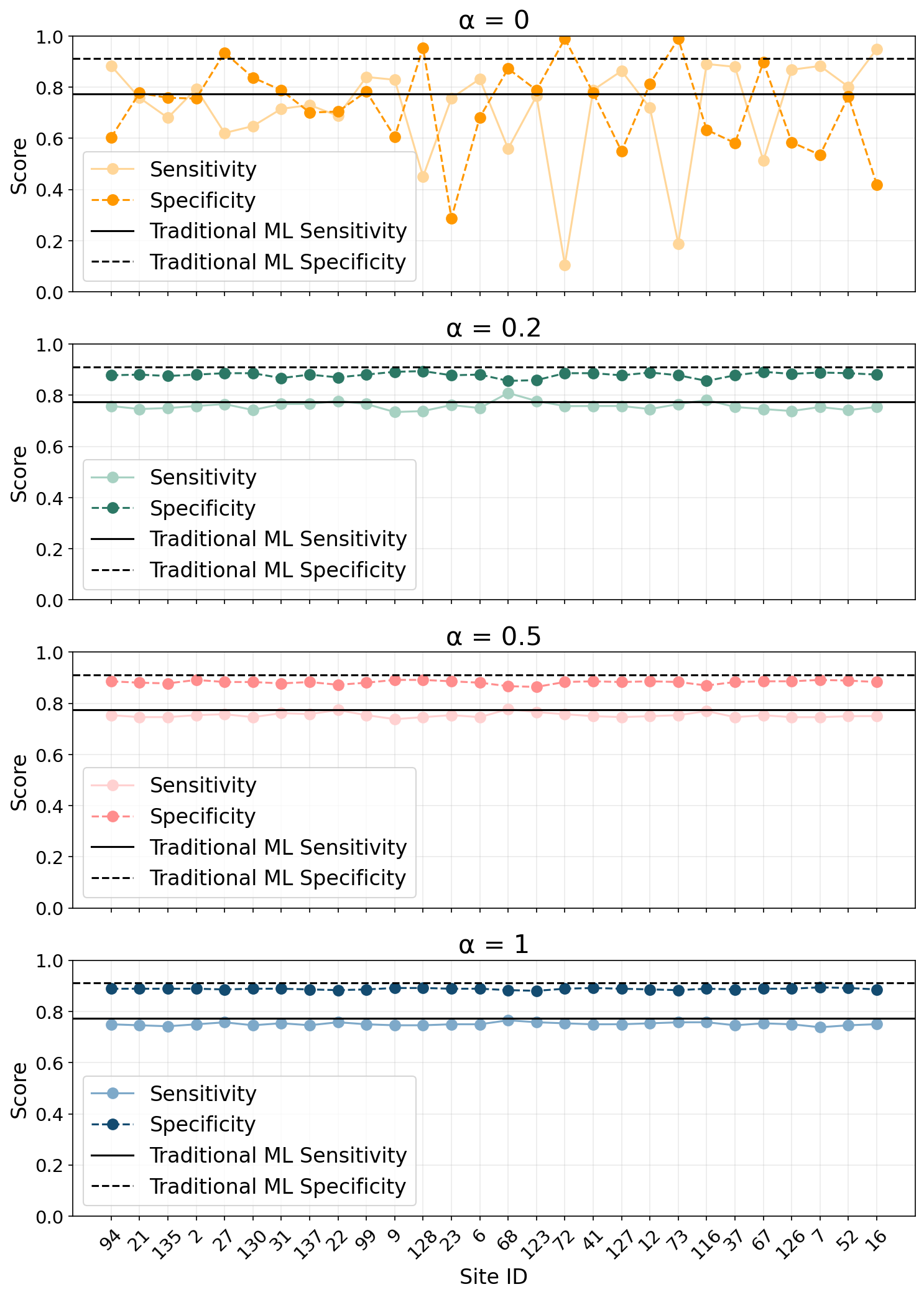}  
        \vfill
    \end{subfigure}%
    \begin{subfigure}[c]{0.5\linewidth}  
        \centering
        \vfill  
        \includegraphics[height=12cm,width=\linewidth]{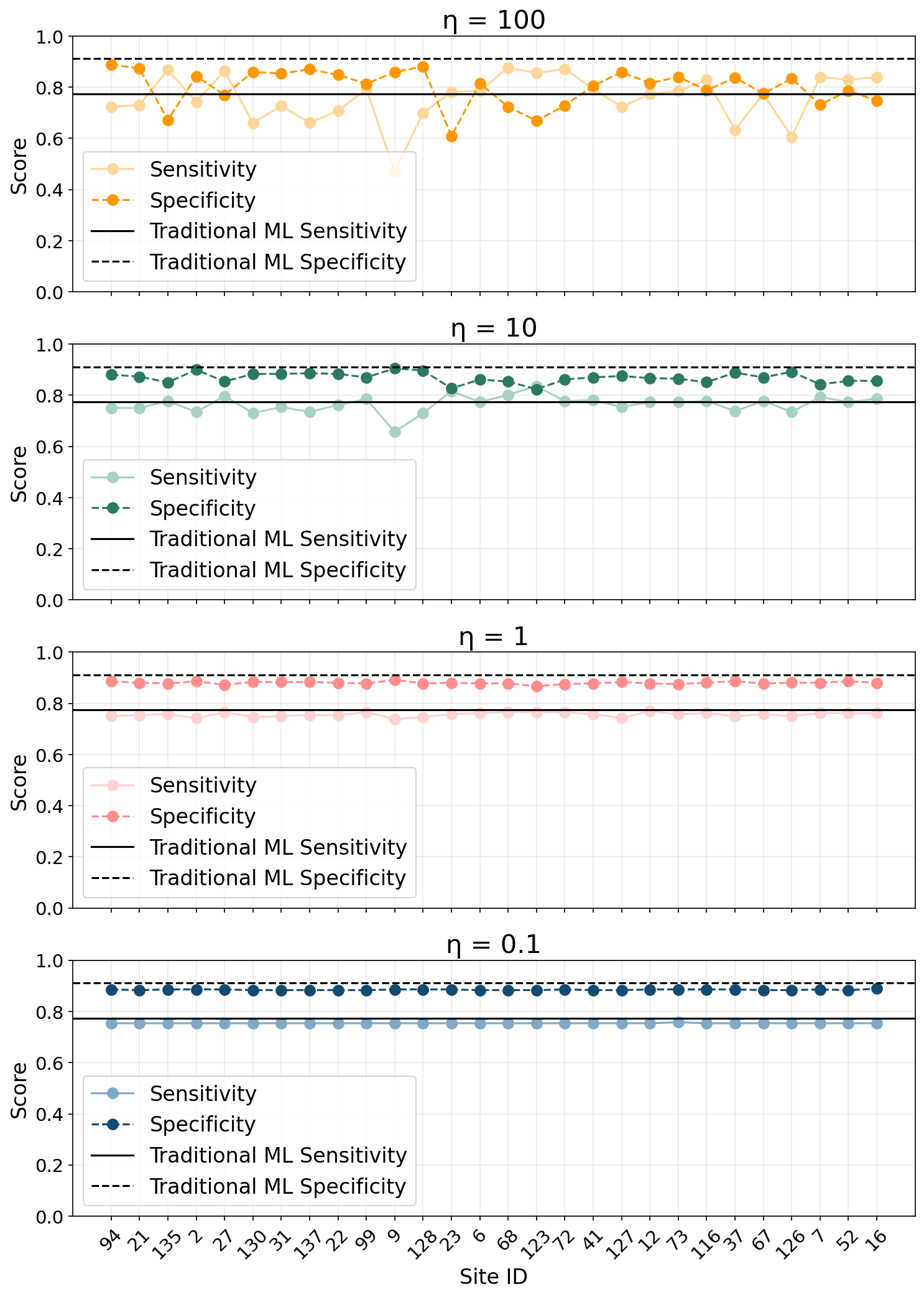}  
        \vfill
        
    \end{subfigure}
    
    \caption{Effect of collaborative parameters on site-specific models for Case 1: predicting cognitive status 3 years from baseline. The left panel illustrates the performance of the P2P architecture using the FedGD algorithm, where the parameter $\alpha \in [0, 1]$ regulates collaboration between local and neighboring nodes. 
The right panel shows the client-server architecture with the FedAvg algorithm, where the parameter $\eta$ controls gradient regularization. 
A higher value of $\alpha$ and a smaller value of $\eta$ constrain local updates to remain closer to those of neighboring nodes or the server, respectively.
    }
    \label{fig:parameterEffect}
\end{figure*}

\begin{figure*}[t]
    \centering
    \begin{subfigure}[t]{0.49\textwidth}
        \centering
        \vspace{0pt} 
        \includegraphics[width=1\linewidth]{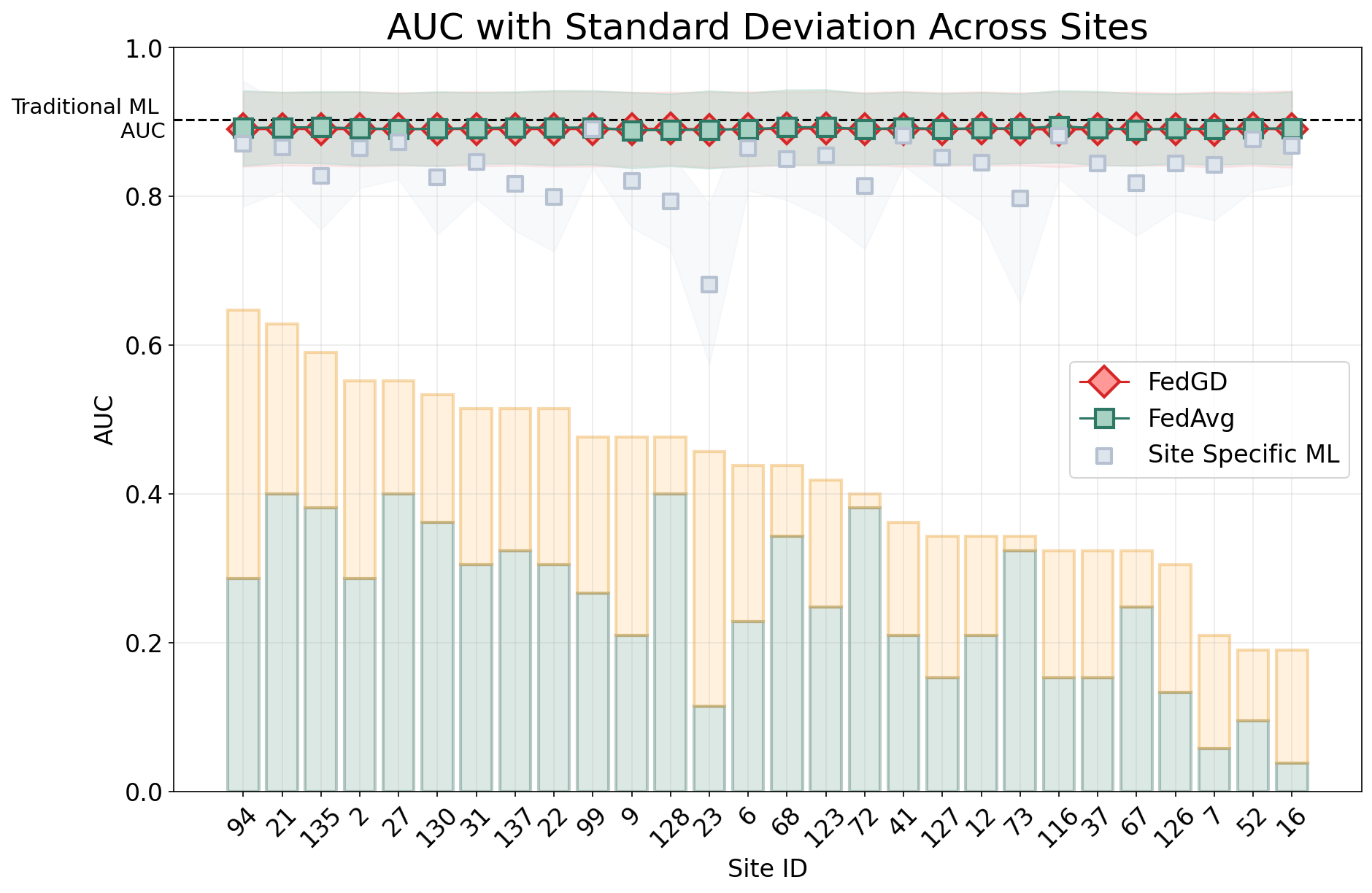}
        \label{fig:3_auc_std}
    \end{subfigure}
    \begin{subfigure}[t]{0.49\textwidth}
        \centering
        \vspace{0pt} 
        \includegraphics[width=1\linewidth]{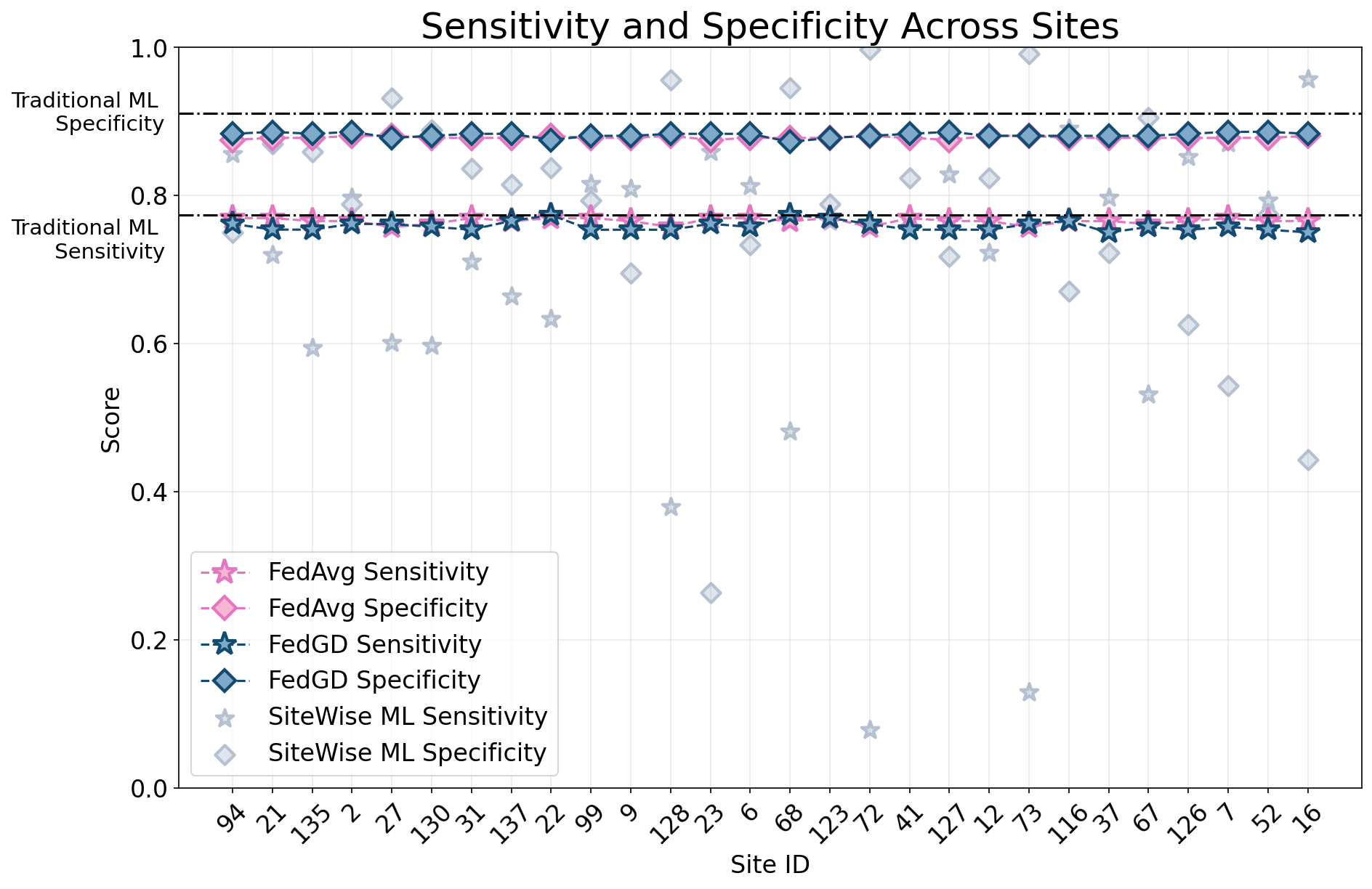}
        \label{fig:3_spec_sens}
    \end{subfigure}
    \caption{Performance metrics comparison of traditional ML (centralized and site-specific) and federated methods (client-server FedAvg and P2P FedGD) for 3-year advanced prediction of MCI to dementia conversion from baseline. The left panel presents AUC ROC scores and depicts the similarity of centralized ML to FL and the performance improvement over site-specific results after collaboration, regardless of participant count and site class distribution. The bar plots in the background represents the participant count where greens represent the number of participants with label 0: sMCI and orange depicts label 1: pMCI. The  right panel visualizes the scores for sensitivity and specificity values. Metrics are averaged across all folds, with parameter values $\alpha = 1$ and $\eta = 0.1$ in the FedGD and FedAvg algorithms.}
    \label{fig:sitewise_metrices_case1}
\end{figure*}

\section{Methods} 
\label{sec:sec_methods} 

We formulate a horizontal FL methodology to predict the conversion from MCI to dementia using a comprehensive set of cognitive and socio-demographic features. This section describes the traditional ML methods (centralized and site-specific) and FL methodologies (with client-server and P2P network architectures).


Let \(\mathcal{D}_k = \{(\mathbf{x}_i, y_i)\}_{i=1}^{n_k}\) represent the local dataset at site \(k\), where \(\mathbf{x}_i \in \mathbb{R}^d\) is the \(d\)-dimensional feature (here \(d = 15\)) vector for the \(i\)-th participant,  \(y_i \in \{0, 1\}\) indicates the participant's conversion status (sMCI or pMCI), and  \(n_k\) is the number of participants at site \(k\).
The total dataset across all sites is \(\mathcal{D} = \bigcup_{k=1}^{K} \mathcal{D}_k\), where \(K\) is the number of participating sites, and the total number of participants is \(n = \sum_{k=1}^{K} n_k\). Each feature vector was normalized by Z-scoring before model training using site-specific or global statistics. 

\subsection{Centralized ML}

We considered the traditional approach, where the entire dataset \(\mathcal{D}\) was transferred to a central location, as a baseline model. In this centralized ML, the data was normalized, and the logistic regression model was trained at the central location 
using the logistic loss function:
\begin{equation}
\mathcal{L}(\mathbf{v}) = -\frac{1}{n} \sum_{i=1}^{n} \left[ y_i \log(h_\mathbf{v}(\mathbf{x}_i)) + (1 - y_i) \log(1 - h_\mathbf{v}(\mathbf{x}_i)) \right],
\label{eq:logisticRegression}
\end{equation}
where \(\mathbf{v} \in \mathbb{R}^d\) is the weight vector of the model and \(h_\mathbf{v}(\mathbf{x}_i) = \frac{1}{1 + e^{-\mathbf{v}^\top \mathbf{x}_i}}\) is the sigmoid function.

\subsection{Site-Specific ML}

In the site-specific ML approach, each site \(k\) independently normalized the local dataset  \(\mathcal{D}_k\) and trained a site-specific logistic regression model  using its private data. The local loss function at site \(k\) is \(\mathcal{L}_k(\mathbf{v})\) is
\begin{equation}
\mathcal{L}_k(\mathbf{v}) = -\frac{1}{n_k} \sum_{i=1}^{n_k} \left[ y_i \log(h_\mathbf{v}(\mathbf{x}_i)) + (1 - y_i) \log(1 - h_\mathbf{v}(\mathbf{x}_i)) \right],
\label{eq:logisticRegression_siteWise}
\end{equation}
This approach reflects independent model training without data sharing or collaboration.

\subsection{FL methodologies}

FL methods utilized client-server and P2P network architectures to train a global model collaboratively. 
Fig. \ref{fig:graphs} visualizes the architectures, where P2P utilized FedGD and client-server used FedAvg algorithm for model training.

\subsubsection{Client-server Architecture with FedAvg}

FedAvg started with the federated data normalization, where each client (node or site) \(k\) shared its local statistics (mean \(\mu_k\) and variance \(\sigma_k^2\)) with a central server computing the global mean \(\mu_{\text{global}}\) and variance \(\sigma_{\text{global}}^2\):

\[
\mu_{\text{global}_j} = \frac{1}{K} \sum_{k=1}^K \mu_{kj}, \quad \sigma_{\text{global}_j}^2 = \frac{1}{K} \sum_{k=1}^K \sigma_{kj}^2,
\]
where 
index $j$ refers to $j$th feature.
Using the global statistics, the local data \(x_{ij}\) at each site \(k\) is standardized as follows:
\[
z_{ij} = \frac{x_{ij} - \mu_{\text{global}_j}}{\sigma_{\text{global}_j}},
\]
where \(z_{ij}\) is the standardized value.
We chose this simple standardization over more complex ones, e.g., by using,  
\(
\quad \sigma_{\text{global}_j}^{2*} = \frac{1}{K} \sum_{k=1}^K \sigma_{kj}^2 + \frac{1}{K} \sum_{k=1}^K (\mu_{kj} - \mu_{\text{global}_j})^2,
\)
as the convergence appeared faster with the simple method. 

The Federated Averaging (FedAvg) algorithm without weighted averaging \cite{mcmahan2017communication} was used for training. For each global iteration \(t\):
\begin{itemize}
    \item The central server broadcasts the current global model parameters, \(\mathbf{w}^{(t)}\), to all sites.
    \item In each site \(k\), local gradient descent updates are performed to minimize the objective:
    \begin{equation}
        \mathbf{w}^{(t)}_k := \arg\min_{\mathbf{v} \in \mathbb{R}^d} \left[ \mathcal{L}_k(\mathbf{v}) + \frac{1}{\eta} \|\mathbf{w}^{(t-1)} - \mathbf{v}\|_2^2 \right],
    \end{equation}
    where \(\mathcal{L}_k(\mathbf{v})\) is the local loss defined in eq. (\ref{eq:logisticRegression_siteWise}) that used the standardized features \(\mathbf{z}_{i}\) and \(\eta > 0\) controls the regularization strength.
    \item The updated parameters \(\mathbf{w}^{(t)}_k\) are sent back to the server, which aggregated them as:
    \begin{equation}
        \mathbf{w}^{(t+1)} = \frac{1}{K} \sum_{k=1}^{K} \mathbf{w}^{(t)}_k.
    \end{equation}
\end{itemize}

\subsubsection{P2P Architecture with FedGD}

In the P2P architecture, we utilized the natural data distribution to formulate the empirical graph, where sites exchanged local statistics with their neighbors to compute neighborhood-wide means and variances, ensuring decentralized normalization. In case of no collaboration ($\alpha = 0$), the data normalization was done independently.

For model training,  we employed the Federated Gradient Descent (FedGD) algorithm for generalized total variation minimization (GTVmin) \cite{jung2024analysis, mcmahan2017communication}. At each global iteration \(t\), site \(k\) updated its parameters as:
\begin{equation}
\mathbf{w}^{(t+1)}_k := \arg\min_{\mathbf{v} \in \mathbb{R}^d} \left[ \mathcal{L}_k(\mathbf{v}) + \alpha \sum_{k' \in N(k)} A_{k,k'} (\mathbf{w}^{(t)}_{k'} - \mathbf{w}^{(t)}_k) \right],
\end{equation}
where
\(\mathcal{L}_k(\mathbf{v})\) is the local loss defined in eq. (\ref{eq:logisticRegression_siteWise}), \(N(k)\) is the set of neighbors of site \(k\), \(A_{k,k'}\) is the weight of the edge between sites \(k\) and \(k'\), and \(\alpha \in [0, 1]\) is a tunable parameter controlling the influence of neighbors.
We used uniform weights (\(A_{k,k'} = 1\)) to ensure equal contributions from all neighbors.

Both FedAvg and FedGD were run for a fixed number of iterations, ensuring convergence and maintaining comparability between the two methods.

\subsection{Implementation and Evaluation}

We used a 10-fold stratified cross-validation procedure for training and evaluation. One fold was reserved for testing, while the remaining folds were used for training. The evaluation was performed using a centralized test set (reserved fold) as: 1) the primary purpose of this study is to compare different FL network architectures to the centralized ML baseline; 2) the participant count at all sites was limited to formulate a site-based test set, both in FL and site-specific ML. Remaining folds were then distributed among sites where stratification ensured consistent class distribution. 

We trained and simulated  ML and FL using a high-performance laptop equipped with a 13th Gen Intel\textsuperscript{\textregistered} Core\texttrademark\ i7-1365U processor featuring 16 cores and 32 GB of RAM. 
The client-server and P2P architectures were simulated using the NetworkX Python package. 
The fixed iteration count for FL algorithms was 3000 global and one local iteration per round. Unless otherwise specified, FedAvg and FedGD hyperparameters were set to \(\eta = 0.1\) and \(\alpha = 1\), respectively.


\begin{figure}[]
    \includegraphics[width=\linewidth]{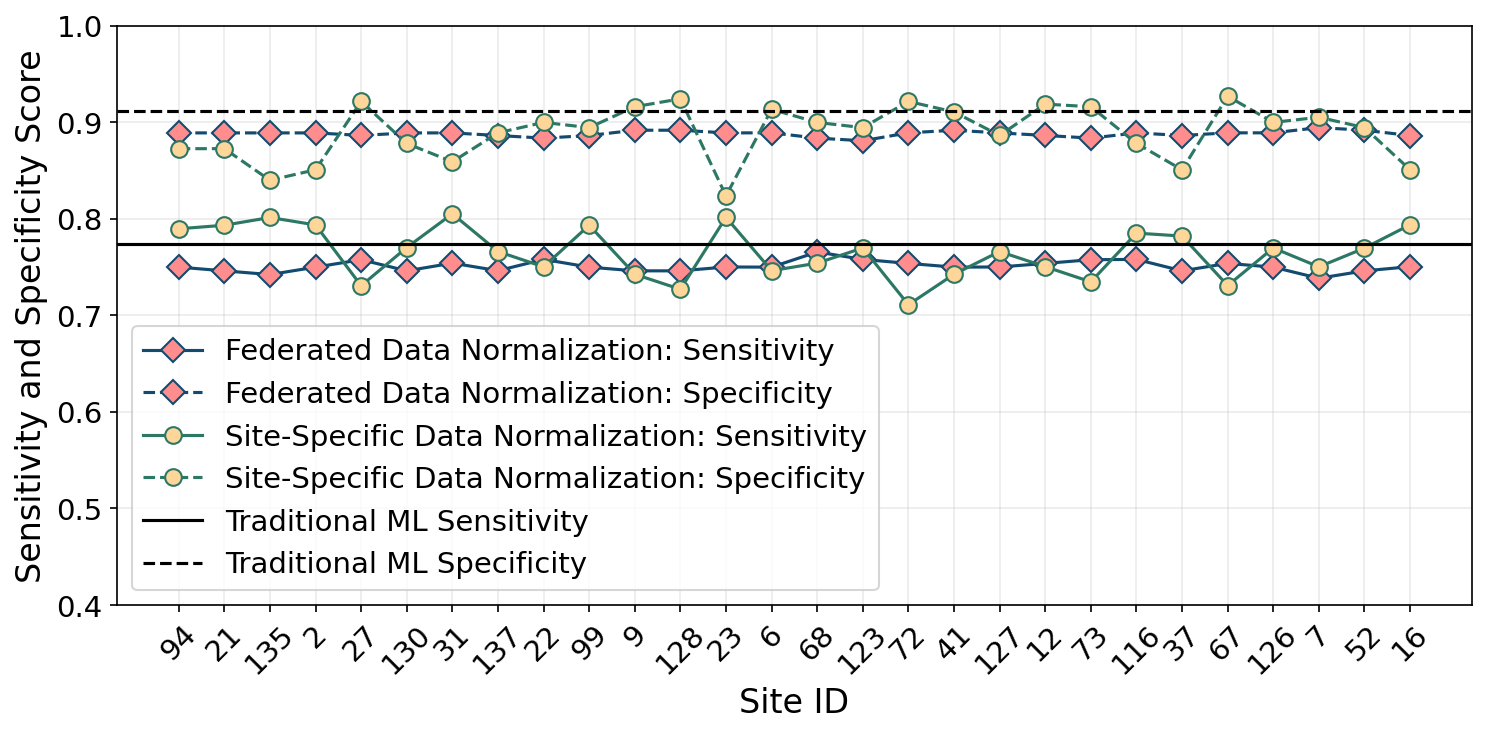}
  \caption{Effect of federated and site-specific data normalization at \(\alpha = 1\) in the FedGD P2P architecture. The scores are represented as the mean of all folds.}
  \label{fig:normalization}
\end{figure}

\section{Results} 
\label{sec:sec_results} 



\subsection{Overall Performance}

Table \ref{tab:logisticRegression} compares the performance of FL methodologies with centralized ML across various evaluation periods. The results indicate an $\approx$ 2\% or less difference between the two methodologies, with slight improvements in ROC AUC and balanced accuracy as the prediction period extended from 3 to 5 years. Sensitivity and specificity scores at cut-off 0.5, became more balanced over time, achieving comparable levels by the 5-year mark. 
These results affirmed that FL can achieve performance equivalent to centralized ML. 

Figure \ref{fig:parameterEffect} further visualizes the effect of collaboration: as the parameter \(\alpha\) approached 1, the local models across sites converged toward similar performance. A comparable trend was observed in the FedAvg algorithm when the gradient regularization parameter \(\eta\) decreased toward 0. 
With the increase in global iterations, there was an interchange between the sensitivity and the specificity scores was observed stabilizing after 3000 iterations. 

\subsection{Site-wise Performance}

Figure \ref{fig:sitewise_metrices_case1} illustrates the site-wise performance and the impact of local parameter sharing on all local models. The mean and standard deviation of ROC AUC scores across sites are shown in left panel, while right panel highlights the FL's sensitivity and specificity scores for all sites compared to the centralized ML and site-specific independent models. These results demonstrate the effectiveness of federated collaboration. Notably, in both FedGD and FedAvg methods, all sites achieved performance levels comparable to centralized ML and far better than site-specific independent models. 

The site-specific analysis reveals noticeable variability in ROC AUC scores when a site operates independently. For instance, Site ID 23, despite having a relatively high participant count, achieved an ROC AUC score of only 0.62 due to unbalanced data. However, through federated collaboration, the score improved significantly to 0.89. Similarly, Figures \ref{fig:sitewise_metrices_case1}  and \ref{fig:parameterEffect} emphasize improved generalizability through FL. Sites 72 and 73, which had only one pMCI participant each, performed similarly to other sites after collaboration. For Site ID 73, the P2P approach improved sensitivity from 0.14 to 0.74 and maintained a specificity of 0.91, compared to its pre-collaboration specificity of 0.99.

\subsection{Data Normalization and Computation Time}

\subsubsection{Collaborative Data Normalization}

The comparison between collaborative and site-specific data normalization demonstrated that the FedGD algorithm achieved higher sensitivity and consistent performance across sites in collaborative settings, as visualized in Fig. \ref{fig:normalization}. Conversely, with site-specific data normalization, it exhibited greater variance in scores. 
The FedAvg algorithm provided a consistent performance on the given dataset in both normalization scenarios.

\subsubsection{Computation Time for Training}


The total computation time for the client-server FL architecture was 0.12 seconds.
In the P2P architecture, the computation time increased with the number of edges per node. For instance, with three edges per node, the simulation required 0.50 seconds. However, with ten edges per node, the time increased to 0.85 seconds.
However, compared to FL, the centralized ML setup, being straightforward, required only 0.0024 seconds for simulation.

\section{Discussion} 
\label{sec:sec_discussion}


Identifying people who experience early cognitive changes that indicate a possible progression to dementia is essential. Achieving this goal requires sufficient diagnostic resources at each clinical site.
To equip a site with a trained predictive model, centralized ML methodology demands the sharing of sensitive information with a central location, raising privacy and data confidentiality risks.
We addressed these issues by training local predictive models distributively using two FL architectures, P2P and client-server. Our federated approach keeps the data local and predicts the progression from MCI to dementia using socio-demographic and cognitive measures. 

 Previous works of FL in research related to dementia \cite{lei2023federated, 9630382, stripelis2021secure, stripelis2023secureprivatefederated, elsersy2023federated, mitrovska2024secure} have mainly considered imaging and blood-based biomarkers. However, these modalities are either invasive or expensive \cite{colliot2023machine, ANSART2021101848, grassi2019novel}.
Cognitive measures are both cost-effective and non-invasive but have not been previously investigated in the FL setting. Regardless of the fact that they provide optimal results alone \cite{turner2023standardized, elkana2015sensitivity} or better when mixed with imaging and other modalities \cite{ANSART2021101848}, to our knowledge, they have yet to be explored in the FL setup.

Our results indicate that socio-demographic and cognitive measures provide a 90\% AUC ROC score for 3 to 5 years follow-up from the baseline for predicting MCI to dementia conversion using centralized ML. FL, both with the client-server and P2P network architecture, provided performance comparable to centralized ML. 
Notably, the federated approach enabled sites such as Site IDs 23, 72, and 73 to improve performance substantially after collaboration, despite imbalanced data or low participant counts.

Recent research suggests that federated data normalization \cite{marchand2022securefedyj, guerraoui2024overcoming} is essential to address high data heterogeneity and variation. In our case, we noticed the effect in the FedGD algorithm; federated data normalization aided in converging local models faster and led to less discrepancy at $\alpha=1$ compared to the site-specific data normalization. 


For healthcare applications, the client-server architecture using the FedAvg algorithm offers a better, computationally efficient, and practical solution to perform distributed analysis and training.
Its simplicity, lower computational demand, and ability to maintain privacy make it a practical solution for distributed datasets. Furthermore, a client-server setup minimizes bureaucratic challenges associated with inter-site agreements, which are common in healthcare collaborations.

Despite the promising results, this study has potential limitations. The evaluation was conducted using simulated networks, which may not fully capture real-world complexities such as hardware heterogeneity, network communication delays, or regulatory constraints. Future work should validate these findings in operational settings. Moreover, the study focused exclusively on horizontal FL with a shared feature space across sites. Real-world scenarios often involve different cognitive tests performed in different countries. 
Expanding this research to address such configurations could further enhance FL's applicability.
Finally, while the study highlighted computational efficiency, it did not address fine tuning of graphical structures and edge weights. Future research could explore optimization techniques for P2P and client-server architectures.


\section{Conclusion}
\label{sec:conclusion}

This study demonstrated that FL can achieve performance equivalent to centralized ML and far better than site-specific ML models, while keeping data private and improving outcomes for sites with limited data. Among the network architectures explored, the client-server, with the FedAvg approach, stood out for its computational efficiency, simplicity, and practicality, making it a compelling choice for real-world applications. These findings pave the way for deploying FL in distributed networks to tackle challenges in sensitive domains such as dementia prediction and beyond. 

\section*{Acknowledgment}
Data collection and sharing for this project was funded by the Alzheimer’s Disease Neuroimaging Initiative (ADNI) (National Institutes of Health Grant U01 AG024904) and DOD ADNI (Department of Defense award number W81XWH-12-2-0012). ADNI is funded by the National Institute on Aging, the National Institute of Biomedical Imaging and Bioengineering, and through generous contributions from the following: AbbVie, Alzheimer’s Association; Alzheimer’s Drug Discovery Foundation; Araclon Biotech; BioClinica, Inc.; Biogen; Bristol-Myers Squibb Company; CereSpir, Inc.; Cogstate; Eisai Inc.; Elan Pharmaceuticals, Inc.; Eli Lilly and Company; EuroImmun; F. Hoffmann-La Roche Ltd and its affiliated company Genentech, Inc.; Fujirebio; GE Healthcare; IXICO Ltd.; Janssen Alzheimer Immunotherapy Research \& Development, LLC.; Johnson \& Johnson Pharmaceutical Research \& Development LLC.; Lumosity; Lundbeck; Merck \& Co., Inc.; Meso Scale Diagnostics, LLC.; NeuroRx Research; Neurotrack Technologies; Novartis Pharmaceuticals Corporation; Pfizer Inc.; Piramal Imaging; Servier; Takeda Pharmaceutical Company; and Transition Therapeutics. The Canadian Institutes of Health Research is providing funds to support ADNI clinical sites in Canada. Private sector contributions are facilitated by the Foundation for the National Institutes of Health (\url{http://www.fnih.org}). The grantee organization is the Northern California Institute for Research and Education, and the study is coordinated by the Alzheimer’s Therapeutic Research Institute at the University of Southern California. ADNI data are disseminated by the Laboratory for Neuro Imaging at the University of Southern California.

\bibliographystyle{IEEEbib}
\bibliography{FL_MCI_D_references}

\end{document}